\documentclass[letterpaper]{article} 
\usepackage{aaai25}  
\usepackage{times}  
\usepackage{helvet}  
\usepackage{courier}  
\usepackage[hyphens]{url}  
\usepackage{graphicx} 
\usepackage{natbib}  
\usepackage{caption} 
\usepackage{algorithm}
\usepackage{algorithmic}
\usepackage{xcolor}
\usepackage{amssymb}
 \usepackage{amsmath}
\usepackage{newfloat}
\usepackage{listings}
\DeclareCaptionStyle{ruled}{labelfont=normalfont,labelsep=colon,strut=off} 
\floatstyle{ruled}
\newfloat{listing}{tb}{lst}{}
\floatname{listing}{Listing}
\newcommand{\commentout}[1]{}
\title{Explainable AI Approach using Near Misses Analysis}
\author {
    Eran Kaufman\textsuperscript{\rm 1},
    Avivit Levy\textsuperscript{\rm 1},
}
\affiliations {
    \textsuperscript{\rm 1}Shenkar Collee of Engineering.Design.Art\\
    erankfmn@gmail.com, avivitlevy@shenkar.ac.il
}

\begin{document}

\maketitle

\begin{abstract}
This paper introduces a novel XAI approach based on near-misses analysis (NMA). This approach reveals a hierarchy of logical 'concepts' inferred from the latent decision-making process of a Neural Network (NN) without delving into its explicit structure. We examined our proposed XAI approach on different network architectures that vary in size and shape (e.g.,\ ResNet, VGG, EfficientNet, MobileNet) on several datasets (ImageNet and CIFAR100). The results demonstrate its usability to reflect NNs latent process of concepts generation. Moreover, our experiments suggest that efficient architectures, which achieve a similar accuracy level with much less neurons may still pay the price of explainability and robustness in terms of concepts generation. We, thus, pave a promising new path for XAI research to follow.
\end{abstract}

%

\section{Introduction}\label{sec:intro}
Explainable Artificial Intelligence (XAI) is an artificial intelligence model incorporated with a set of tools and frameworks to help users understand and interpret their predictions and decisions. Many AI models, especially Neural Networks (NN), are usually constructed as a "blackbox" model, where often even its designers cannot explain why a specific decision is arrived by the model. 

There are several advantages for an XAI over such "blackbox" AI models.
The first is for applications with high stakes decision impact, such as medicine procedures or large financial transactions
\cite{CausabilityMed,electronics8080832,SurveyMedical}, where understanding of the decision process is important for establishing confidence in the decision. The second is in fields where the decision is not made in isolation, but as part of a larger group of behaviors. For example, an applicant who is denied credit but wishes to understand how his credit can be improved. In addition, there may be external social concerns, such as avoiding biased decisions such as gender or minority discrimination
\citep{DetectingBias,DBLP:journals/corr/abs-1711-01134}.

XAI shows improved performance in both accuracy and robustness when
compared to traditional Deep Neural Networks (DNN)\commentout{ even in traditional settings such as vision}.
For example, \cite{Galloway2018PredictingAE} showed that structured learning is more robust against adversarial attacks than
classical methods. 

Hierarchical learning or concept learning is a form of XAI, which imposes a formal hierarchy on a decision so that a final decision is a path in a hierarchical tree of hypernyms, which expresses increasingly specific characterization, as in the following path: entity-organizm-animal-vertebrate-mammal-cat. Hierarchical learning methods are believed to be superior than traditional methods in tasks such as 'few-shot' learning \cite{DBLP:conf/iclr/MaoGKTW19}.


\paragraph{Paper Contribution.}
We introduce a novel XAI approach based on near-miss analysis (NMA). This approach reveals a hierarchy of logical 'concepts' inferred from the latent decision-making process of a NN without delving into its explicit structure. For instance, one type of network might develop the concept of 'fruits', recognizing an orange by first categorizing it as a fruit, while another model might form the concept of 'oval things,' recognizing an orange through the hierarchical steps: 'oval things'-'orange'. Such differences can arise from variations in the model architecture such as the number of layers or from a difference in the training data. 

NMA can lead to insights on preferring certain architectures for specific tasks. Issues and obstacles regarding the training data can also be uncovered using this approach since a model's performance is shaped by both its design and the data it is trained on.
Thus, our approach can aid in understanding and debugging NNs.\commentout{ For example, if a model erroneously links dogs with cars due to encountering images of dogs in cars during the training procedure, this flawed linkage can be revealed through the NMA. The model designer can then decide to retrain the model without these images in order to improve its accuracy.

In addition, our approach supports both 'black box' explanations as well as 'white box' network dissection methods. By examining the model layer by layer, we can identify where new concepts are formed and in which neurons. NMA also enhances adversarial attack strategies by pinpointing which labels are closely related within the same concepts, thus, can aid developing countermeasures against them.
{\color{red}\\ AL: We promise things that are not demonstrated in this paper. We better mention as contributions only what we actually demonstrate.}}

\section{Related Work}\label{sec:related}
XAI broad field can be categorized in various ways, such as model specific vs.~model agnostic, meaning whether this explanation was constructed specifically and intertwined with a specific model or a family of models or not.
XAI methods can also be categorized as designed 
 vs.~ad-hoc, meaning whether the explanation is constructed during the learning phase or in retrospect.
Further, an explanation may be {\em local}, applying to the respective decisions for each
data point, or {\em global}, by applying to the overall decision process.
 
Many model-specific methods were designed for the field of computer vision.
These include ad-hoc gradient methods, such as guided back propagation \citep{DBLP:journals/corr/SpringenbergDBR14}, 
integrated gradients \citep{DBLP:conf/icml/SundararajanTY17},
SmoothGrad saliency maps \citep{DBLP:journals/corr/SmilkovTKVW17},
contrastive gradient-based saliency maps \citep{DBLP:journals/corr/SimonyanVZ13},
excitation back propagation \citep{DBLP:journals/ijcv/ZhangBLBSS18} 
and Grad-CAM \citep{DBLP:journals/ijcv/SelvarajuCDVPB20}.
Other types of ad-hoc model specific explanation method include knowledge distillation,
which compresses a complex model into a simpler one \citep{DBLP:conf/iclr/PolinoPA18},
 by using tree regularization \citep{DBLP:conf/aaai/WuHPZ0D18,DBLP:journals/corr/HintonVD15} 
or distilling a NN into a soft decision tree \citep{DBLP:conf/aiia/FrosstH17}.

Other works in this field include explainable methods for graph Convolutional Neural Networks (CNN) \citep{DBLP:conf/cvpr/PopeKRMH19}, interpretable and fine-grained visual explanations for CNNs \citep{DBLP:conf/cvpr/WagnerKGHWB19}, interpreting CNNs via Decision Trees
  \citep{DBLP:conf/cvpr/ZhangYMW19}, and learning to explain with complementary examples \citep{DBLP:conf/cvpr/KanehiraH19}.
   
Model agnostic XAI gives a model-independent (blackbox) explanation that is ad-hoc. Among the most famous of these methods are feature importance \cite{DBLP:conf/scisisis/LeeKJKKK20}, Shapley values estimation (SHAP) \citep{DBLP:conf/nips/LundbergL17} and Local Interpretable Model-agnostic Explanation (LIME) \citep{DBLP:conf/aaai/Ribeiro0G18}. By running a set of sliding windows or using local feature descriptors (such as SIFT), one can extract features from an image and by using the model agnostic explanation assign importance to different parts of the image. 

\commentout{One major drawback of the above methods is that they have a non-deterministic component, which leads to low stability of the produced explanation. That is, repeating the explanation generation for the same model may result in different explanations. For example, LIME and SHAP include a random data sampling process and feature importance relays on random shuffling. Further, all above methods contain a linear approximation component that assumes independence between data features, which is often an invalid assumption.}

Many methods have been developed in the field of hierarchical learning. Among these is the aforementioned structured learning 
\citep{Bui2018NeuralGL,Stretcu2019GraphAM,Gopalan2020NeuralSL,DBLP:conf/wsdm/GopalanJMFHLPYF21}.
These are model-specific XAIs. 
In structure learning, the features of a training sample are fed into the model along with some predefined or learned structure.
For example, the pixels of an image are fed along with a graph representing the similarity among samples. The input structure is used to regularize the training of the NN. Each training sample is augmented to include its neighbor information from the graph,
so both the original training samples and their neighbors are fed into the NN. The difference between the embedding of the sample and its neighbors is calculated and augmented into the final loss function as a regularization parameter. Adding this regularization term allows the NN to learn to maintain similarity between the sample and its neighbours of the same ''concept'' \citep{DBLP:journals/corr/GoodfellowSS14,Galloway2018PredictingAE,Liu2021ADVM211I}.

There are some limitations to the feature-based approach. First, features are not necessarily user-friendly in terms of interpretability. For example, the importance of a single pixel in an image usually does not convey much meaningful interpretation. Second, the expressiveness of a feature-based explanation is constrained by features number.

The concept-based approach addresses both limitations mentioned above. A concept can be any abstraction, such as a color, an object, or even an idea. Given any user-defined concept, although a NN might not be explicitly trained with it, a concept-based approach can detect that this concept is embedded within the latent space learned by the network. In other words, the concept-based approach can generate explanations that are not limited by the feature space of a NN. A representative of this approach is the Testing with Concept Activation Vectors (TCAV)\cite{DBLP:conf/icml/KimWGCWVS18}. For any given concept, TCAV measures the extent of its influence on the model’s prediction for a certain class. For example, TCAV can answer questions such as how the concept of 'striped' influences a model classifying an image as a 'zebra'. Since TCAV describes the relationship between a concept and a class, instead of explaining a single prediction, it provides useful global interpretation for a model’s overall behavior.

Our NMA-based approach steps forward the idea of using concepts in XAI by exploring whether concepts are learned in a network latent features; not only simple visual concepts, such as 'striped' or 'oval', but also more subtle and complex concepts, such as 'furniture' or 'animals'.

\section {Method}\label{sec:method}  
\paragraph{Near Misses Analysis (NMA).}
Our method is based on an analysis of the probability vector resulting from an image classification by a given NN. Given the set $\{X,y\}$, where $X \in \mathbb{R}^n$ is the data and $y \in \mathbb{R}$ is the label, NNs trained on some probabilistic loss function return a vector, where for label $i$ we have the probability $\mathbb{P}(y=i \mid X=x)$. The label with the highest probability in this probability vector is chosen to be the classification result for this image.

Given the decreasingly sorted probability vector $\mathbb{P}$ and a constant $K$, we define the \emph{$K$-near misses} to be the the top $K$ labels in the sorted $\mathbb{P}$. In addition, given a threshold $t$, we define the \emph{$t$-cutoff-near misses} to be the labels $i$ in the sorted $\mathbb{P}$ having $\mathbb{P}(y=i \mid X=x)\geq t$.

\begin{figure}[t]
    \centering
    \includegraphics[width=\linewidth]{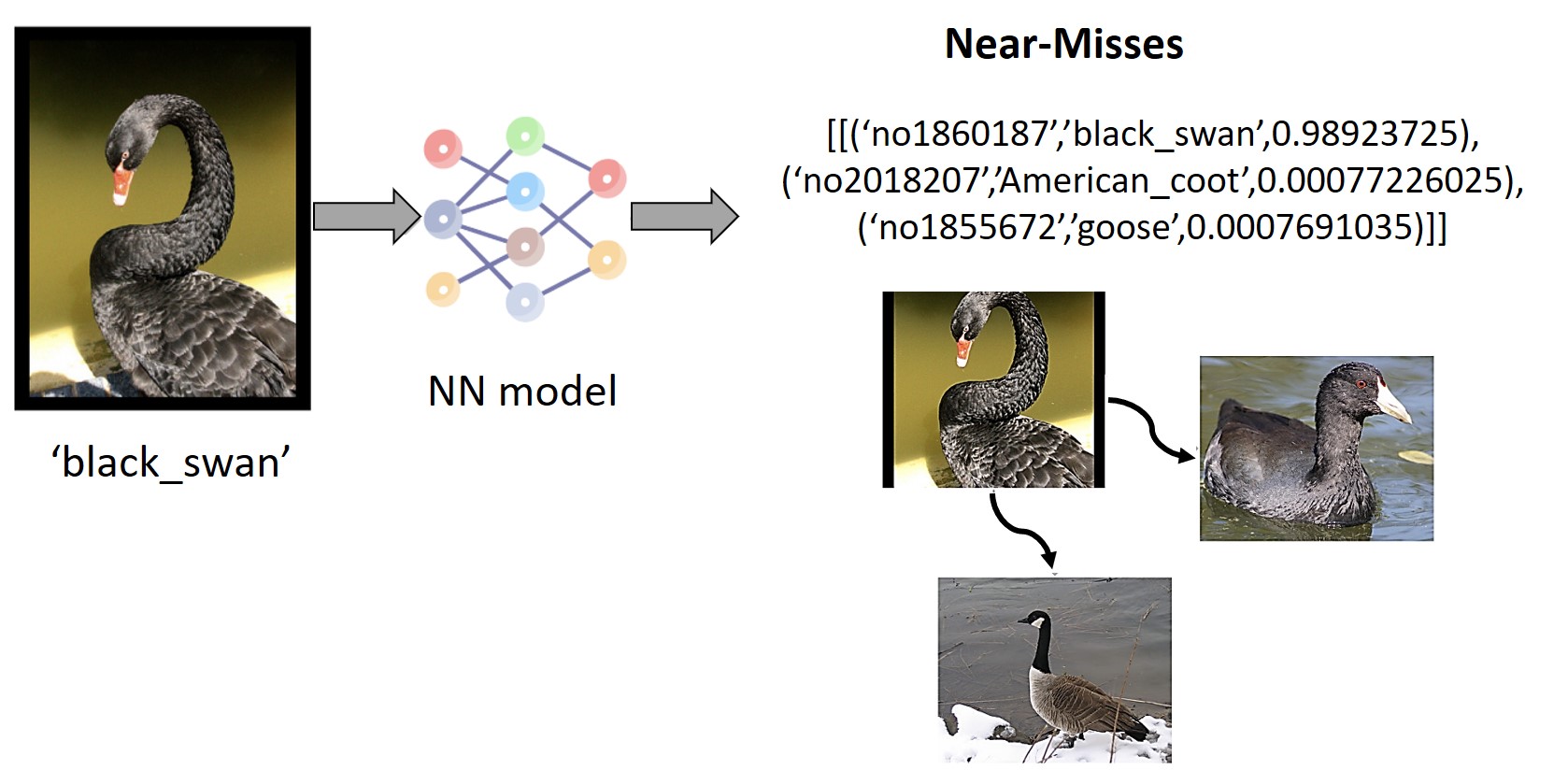}
    \caption{Near-misses for an image classification}
    \label{fig:NearMisses}
\end{figure}

Analysing the near misses in the probability vector reveals connections which the given model makes among the labels and provides a glance to its latent decision-making process. Figure~\ref{fig:NearMisses} shows an example of NMA of a black swan image classification by MobileNetV2. The image of label 'black swan' is correctly classified with the highest probability in the probability vector. Nevertheless, the next two top probabilities are of the labels 'American coot' and 'goose', which have conceptual similarity to 'black swan' as all three are water birds.

We aim to create a hierarchical clustering over the labels on which the model was trained on in order to create ``abstract concepts'' from these labels. To this end, we first build a graph representing the connections revealed between labels in the test set near-misses probability vectors analysis. We then build a hierarchical clustering of the labels over this graph. Specifically, our method consists of the phases:
\begin{enumerate}
\item Connections graph construction (pre-processing): Given a model, perform NMA on the probability vector of each test image classification to create a connections graph.
\item Hierarchical clustering construction (pre-processing): Build a hierarchical clustering given the created connections graph. In this phase, a dendogram describing the hierarchy of the label clusters is produced as \emph{a visual explanation} of the model's decision-process. 
\item Human-readable concepts generation (pre-processing): Generate human-readable concepts for the labels in each cluster of the hierarchy.
\item Query explanation: Given a query image a model and the explainability method described above, create \emph{a verbal explanation} for the classification model for this image. 
\end{enumerate}
We next explain each of these phases in detail.

\subsection{Connections Graph Construction}
Given a model $M$ trained for image classification, we analyze its decision-making process by querying $M$ on each image from a test set, perform NMA and create a weighted connections graph $G=(V,E)$ with weight function $w:E\rightarrow \mathbb{R}$ as follows. 

Let ${\cal X}=\{X,y\}$ be a set and let $l(X,y)$ be the empirical loss calculated by the model $M$. 
For every $(X,i)\in {\cal X}$, i.e.\ the set of all images with label $i$ with cardinality $N$, we denote
$p_{i,j}=\frac{1}{N}\sum_{(X,i)\in {\cal X}}\mathbb{P}(y=j \mid X=x)= E_{(X,i)}(l(x,y)| y=j) $.
$E_{i}(l(x,y)| y=j)$ is the average probability (loss) in which images belonging to label $i$ can be mistaken to be of label $j$, while $E_{i}(l(x,y))$ is defined to be the overall expected loss of all images not belonging to label $i$ 
to be mistaken to be $i$.
Since $p_{i,j}$ acts as a measure of similarity between labels (the higher the probability of erring between labels, the stronger the connection), we take the complementary probability $1-p_{i,j}$ as a measure of dissimilarity between labels. 

For each label $i$, we create a node $v_i\in V$. In addition, for every pair of labels $i,j$, we create an (undirected) edge $(v_i,v_j)\in E$ with weight $w(v_i,v_j)$ as follows:
\begin{displaymath} w(v_i,v_j)=  \left\{ \begin{array}{ll}
     1-p_{i,j}, & \hbox{$p_{i,j} \geq t$};\\
     {\infty}, & \hbox{otherwise}\\
     \end{array}
  \right.
\end{displaymath}
In other words, the weight of an edge $(v_i,v_j)$ reflects the accumulated (joint) \emph{dissimilarity}  between the labels $i$ and $j$ revealed in the test set NMA, where pairs of vertices that had no near-misses in the test set have dissimilarity $\infty$.

\subsection{Hierarchical Clustering Construction}
Given a connection graph $G=(V,E)$ described above we build a hierarchical clustering over $G$ as follows. Compute a shortest-paths-metric by applying Floyd-Warshall algorithm for finding all-pairs-shortest-paths in a given graph. Since the connections graph is undirected, the distances matrix output of this algorithm is a metric. Then, run a hierarchical clustering algorithm on the connections graph based on the shortest-paths-metric. 

A dendogram describing the created hierarchical clustering is produced. This dendogram supplies \emph{a visual explanation} -- a high-level view of the classification decision-making process of the given image classification model for any chosen set of labels. We demonstrate its use in Section~\ref{sec:results}.

\subsection{Human-Readable Concepts Generation}
In addition to the dendogram-based visual explanation, we aim to supply also a verbal explanation for a specific image classification process of the model $M$. Such a verbal explanation follows the path in the given created hierarchical clustering from the classified label to the root and specifies the encountered clusters names while skipping unnamed nodes or similar named nodes in the dendogram. We employ a human-readable concept generation procedure for each cluster, thus, reporting the encountered clusters names in this path acts as a verbal explanation of the model's decision. 

\paragraph{Automatic Concept Naming.} We exploit WordNet lexical dataset for defining an automatic human-readable concept naming procedure as follows.
At each stage of the dendrogram a group is formed. The automatic procedure scans WordNet hypernyms and finds the lowest common hypernym for all members in the group. For example, if the group \{'bed', 'chair'\} is formed, then for the label 'chair', WordNet returns the linkage 'chair'-'furniture'-'entity', and for the word 'bed', WordNet returns the linkage 'bed'-'furniture'-'entity'. Thus, the lowest common hypernym is 'furniture' for this set. If, for example, at the next step of the hierarchical clustering procedure
the label 'armchair' is added to this group with the linkage 'armchair'-'furniture'-'entity', the group will still be given the lowest common hypernym, which is 'furniture'. Nevertheless, the description 'furniture' will not appear twice at the verbal explanatory stage. 
If at the next stage of the hierarchical clustering procedure the group \{'chair','bed','armchair','monkey'\} is formed with 'monkey'
having the linkage 'monkey'-'primate'-'mammal'-'entity', then the entire group will be given the concept name 'entity'. Thus, the query 'chair' in this setting will return the explanation by reading the dendrogram bottom-up and specify the names of the encountered clusters groups: 'chair is apart of the concept furniture which is a part of the concept entity'.

\paragraph{Human-in-the-Loop AI.}
While the automatic concept naming procedure using WordNet provides human-readable concepts, in some cases, the provided names do not well express the shared property by the labels in a given cluster. For example, the name 'entity', which is occasionally given by the automatic concept naming procedure that exploits WordNet, is indeed possible for describing the set of labels: \{'broccoli','head cabbage',  'orange', 'Granny Smith', 'zucchini', 'cauliflower', 'lemon', 'fig'\}, however, the shared property of this group is better described as: 'food'. Therefore, our XAI approach supports a user-aided fine-tuning for the automatic concept naming procedure via a user-interface (UI). The UI enables the user to scan all clusters in a given created hierarchical clustering and manually refine the WordNet automatically suggested name. This stage concludes the human-aid at the explanation 'training' part. 

\subsection{Query Explanation}
When the human-readable concepts are generated for each cluster in a created hierarchical clustering for a given NN, we can support the following query verbal explanation procedure. Given a query image, it is classified by the NN to a decided label. A verbal explanation to the user of this decision process is a sequence of English sentences of the form: ``A $\cal{X}$ is part of the concept $\cal{Y}$'' followed by the labels in the cluster with name $\cal{Y}$. Such a sentence is created for each cluster encountered in the given created hierarchical clustering path from the classified label to the root cluster. 

\paragraph{Explanatory Measurement.}
In order to quantify the query explanation performance for a given model we devised a scoring method based on a comparison between the machine generated and a human annotated explanations. As mentioned, the resulting NMA hierarchical clustering can be transformed into a human explainable sentence given human annotated hypernyms, such that each group formed by the NMA hierarchical clustering at each level gets the group lowest common hypernym. Thus, we can measure how many joint concepts the machine generated explanation and a human explanation have in common.

In order to understand the intuition behind this scoring method consider the following example. Assume that the explanation for the label 'apple' based on the WordNet hierarchy is 'apple-fruit-food-entity' while the machine generated explanation for label 'apple' is 'apple-fruit-entity'. Then, the the machine explanation score of 'apple' is $1$ (fruit) over $2$ (fruit,food - the root (entity) and leaf (apple) are not counted). Assuming that the concept 'fruit' in this example is achieved from clustering apples with oranges, which also has the hypernym vector 'orange-fruit-food-entity', thus orange receives the machine explanation 'orange-fruit-entity' and the machine explanation score for 'orange' is also $\frac{1}{2}$. In addition, if apples and oranges are clustered in a hierarchical level with a subset of the vegetables, their explanation is more compatible with human reasoning (by this exemplary human annotation) and receives a higher score. If, for example, grapes are not clustered with other fruits, their machine generated explanation would be 'grape-entity', which receives a null score. 

A model score is the average score for all labels. We can, thus, compare models trained on the same dataset. Formally, let $M$ be a model trained on the set of labels $\{l_1, \dots.l_N\}$. Let $S_i$ be the set of concepts in the explanation for label $l_i$ (without root and leaf) given by human annotation and let $U_i$ be the set of concepts in the machine generated explanation given by model $M$ to label $l_i$. Then, the score function $\phi$ is:
\begin{equation}\label{eq:score1}
         \phi(l_i)=\frac{|S_i \cap U_i|}{|S_i|} \\
\end{equation}
and
\begin{equation}\label{eq:score2}
         \phi(M)=\frac{1}{N}\sum_i \phi(l_i)  
\end{equation}
\section{Experiments}
In order to test our proposed XAI approach, we examined different network architectures (e.g.,\ ResNet, VGG, EfficientNet, MobileNet) on the ImageNet and CIFAR100 benchmark datasets.

\paragraph{The Chosen Datasets.} ImageNet contains 1.6 million images from 1000 classes that focus on objects. ImageNet is organized according to the WordNet hierarchy, in which each node of the hierarchy is depicted by hundreds and thousands of images. WordNet is a large lexical database which links more general synsets (synonymous sets), such as \{'furniture', 'piece of furniture'\} to increasingly specific ones, such as 'bed'. Thus, WordNet states that the concept 'furniture' includes 'bed'; conversely, 'bed' belong to the concept 'furniture'. 
\commentout{{\color{blue}All noun hierarchies are ultimately linked to the root node 'entity'. Hyponymy relation is transitive, i.e.,\ if an 'armchair' is a kind of 'chair' and a 'chair' is a kind of 'furniture', then an 'armchair' is a kind of 'furniture'. - consider removing}}
The experiments on our proposed XAI approach exploit this connection between the WordNet and ImageNet datasets for the query verbal explanation analysis.

\commentout{ Places205 and Places365 contain 2.4 million / 1.6 million images from 205 / 365 different scenes. }CIFAR100 data set has 100 classes containing 600 images each. There are 500 training images and 100 testing images per class. The 100 classes in the CIFAR100 are grouped into 20 superclasses. Each image comes with a "fine" label (the class to which it belongs) and a "coarse" label (the superclass to which it belongs).

\paragraph{The Chosen NN Models.} We tested several different networks which vary in size and shape. The experimented NNs presented in this paper are: ResNet50, EfficientNetB0, VGG16, MobileNetV2.\footnote{Due to pages limit, additional experimented NNs visual explanations results appear in the supplementary material.} Table~\ref{tab:Models} summarizes the properties of these NNs \footnote{https://keras.io/api/applications}. Table~\ref{tab:Models} shows that they achieve a similar Top-5 accuracy, however, VGG has extremely large size, while EfficientNet and MobileNEet are relatively small. On the other hand, VGG is relatively shallow, while the other models are much deeper. ResNet is both a deep NN as well as relatively complex, regarding its learnt parameters-set size. Our premise was that more complex networks will be able to 'think' in more complex and subtle concepts. 

\begin{table}[t]
\centering
\resizebox{.95\columnwidth}{!}{
\begin{tabular}{|c|c|c|c|c|}
\hline
    Model & Size & Top-5 & Parameters & Depth\\
     & (MB) & Accuracy &  & \\
\hline
    ResNet50 & 98 & 92.1\% & 25.6M & 107\\
    EfficientNetB0 & 29 & 93.3\% & 5.3M & 132\\
    VGG16 & 528 & 90.1\% & 138.4M & 16\\
    MobileNetV2 & 14 & 90.1\% & 3.5M & 105\\
\hline
\end{tabular}}
\caption{Experimented NNs properties summary}
\label{tab:Models}
\end{table}

\paragraph{Implementation.}
We implemented a UI for researchers. For the implementation $K$-near-misses are used with $K=3$ and $t$-cutoff with $t=10^-6$. The are two operational modes for developers and users: 'naming' and 'query'. By pressing the 'naming' tab the user receives a dendogram describing the hierarchical clustering created for the pre-processed chosen model and dataset. The clusters are presented by specifying the list of labels included in this cluster, while 'hoovering' on a label in this list presents an image of this label. The user suggestion replaces the previous concept name for this cluster. By pressing the 'query' tab the user can upload an image, then \commentout{and receives the human-readable concepts that were generated for each cluster in a given created hierarchical clustering for a given NN. We support the following query verbal explanation procedure. Given a query image,} it is classified by the NN to a decided label and the classification verbal explanation is presented.\footnote{The code appears here: https://anonymous.4open.science/r/XAI-40AD/README.md  and a demonstrative video appears here: https://www.youtube.com/watch?v=juUMxptjMPI.}

\commentout{ in a sequence of English sentences of the form: ``A $\cal{X}$ is part of the concept $\cal{Y}$'' followed by the labels in the cluster with name $\cal{Y}$, where 'hoovering' on a label presents its image. Such a sentence is created for each cluster encountered in the given created hierarchical clustering path from the classified label to the cluster in hierarchy root.}

\paragraph{Experiments Course.} \commentout{All experimented NNs were trained on \emph{the same dataset training set}.}The \emph{same dataset of images} was used for creating all experimented NNs explanations. 

For each experimented NN the following experiment process was conducted. For each correctly classified test image, a NMA was performed and the appropriate weighted edge was updated in the connections graph. When the pre-processing phase of connections graph construction ended, the pre-processing hierarchical clustering phase was performed. Then, a visual explanation in the form of a dendogram was created for this model and dataset.

\paragraph{ImageNet Test Sample.} The test set used for creating NNs explanations on ImageNet is the miniImageNet benchmark dataset, which is used by many in few-shot learning research \cite{DBLP:journals/corr/abs-2310-10971,DBLP:conf/iclr/Poulakakis-Daktylidis24}. This relatively small dataset resulted in a surprising expressiveness of the resulting explanations.

Since we cannot possibly show the results for all 1000 Imagenet labels, we chose a subset of 41 labels, which can reasonably demonstrate the principles by which the models create new concepts. These labels were taken from 4 different demonstrative concepts: animals, kitchen utensils, vehicles , plants, chosen so that they can demonstrate immediate semantic concepts as well as more complex subtle ones. For example, we chose a few types of cats which are expected to be clustered together as 'cats'. The same was done for dogs and both are expected to be clustered as 'house animals'
and along with several other types of primates should be clustered as 'animals' according to a human explanation.
On the other hand, we took vehicles which are far away in the semantic tree from animals. From this branch we took a few types of automobiles, ships and other water vehicles as well as avionic vehicles, in order to examine whether some common visual latent features extracted by the network could cluster all of them together as 'transportation vehicles'.

\section{Results and Discussion}\label{sec:results}
\begin{figure}[ht]
    \centering
    \includegraphics[width=\linewidth]{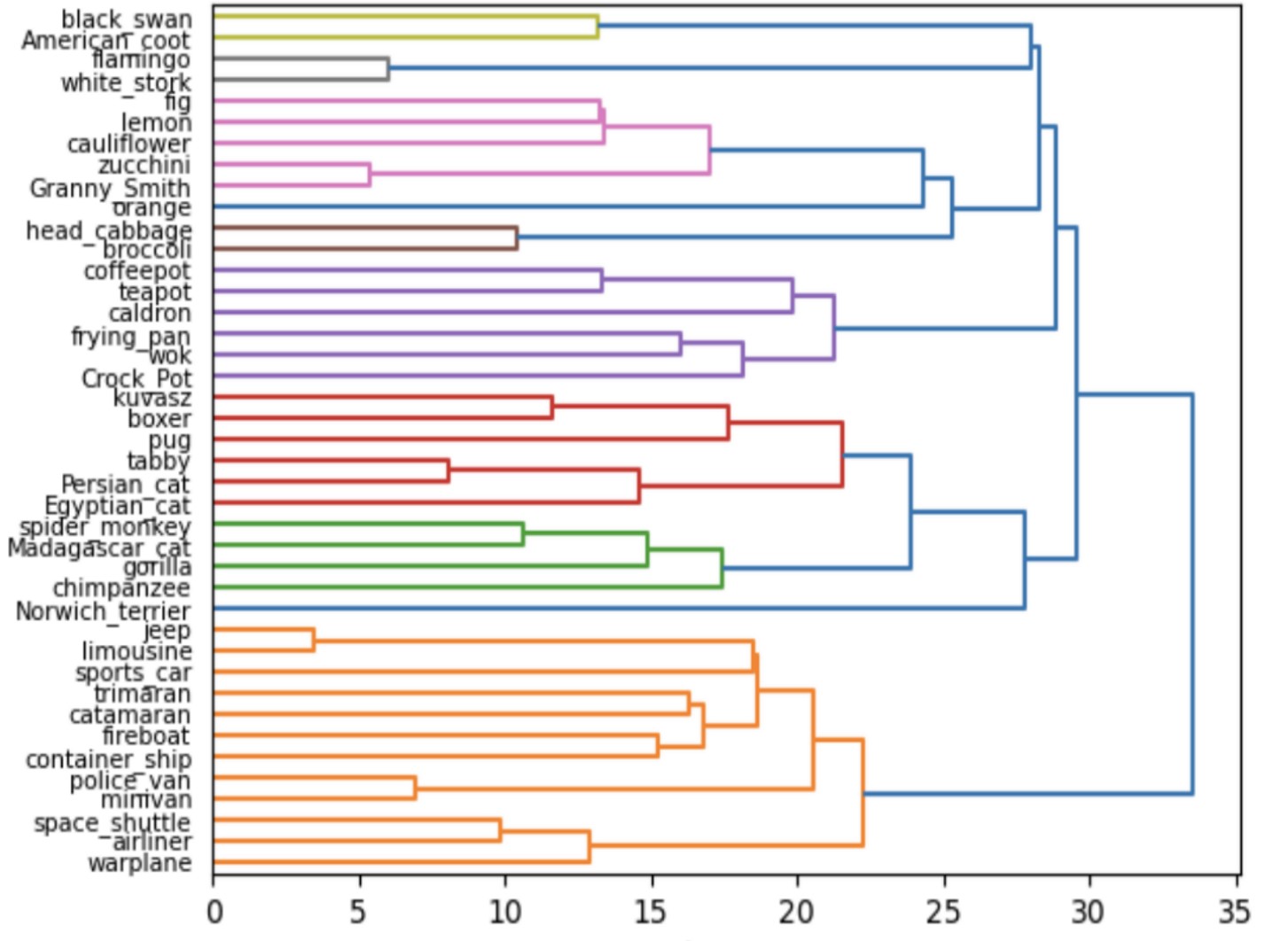}
    \caption{Visual explanation of ResNet on ImageNet.}
    \label{fig:ResNetImgNet}
\end{figure}

\begin{figure}[ht]
    \centering
    \includegraphics[width=\linewidth]{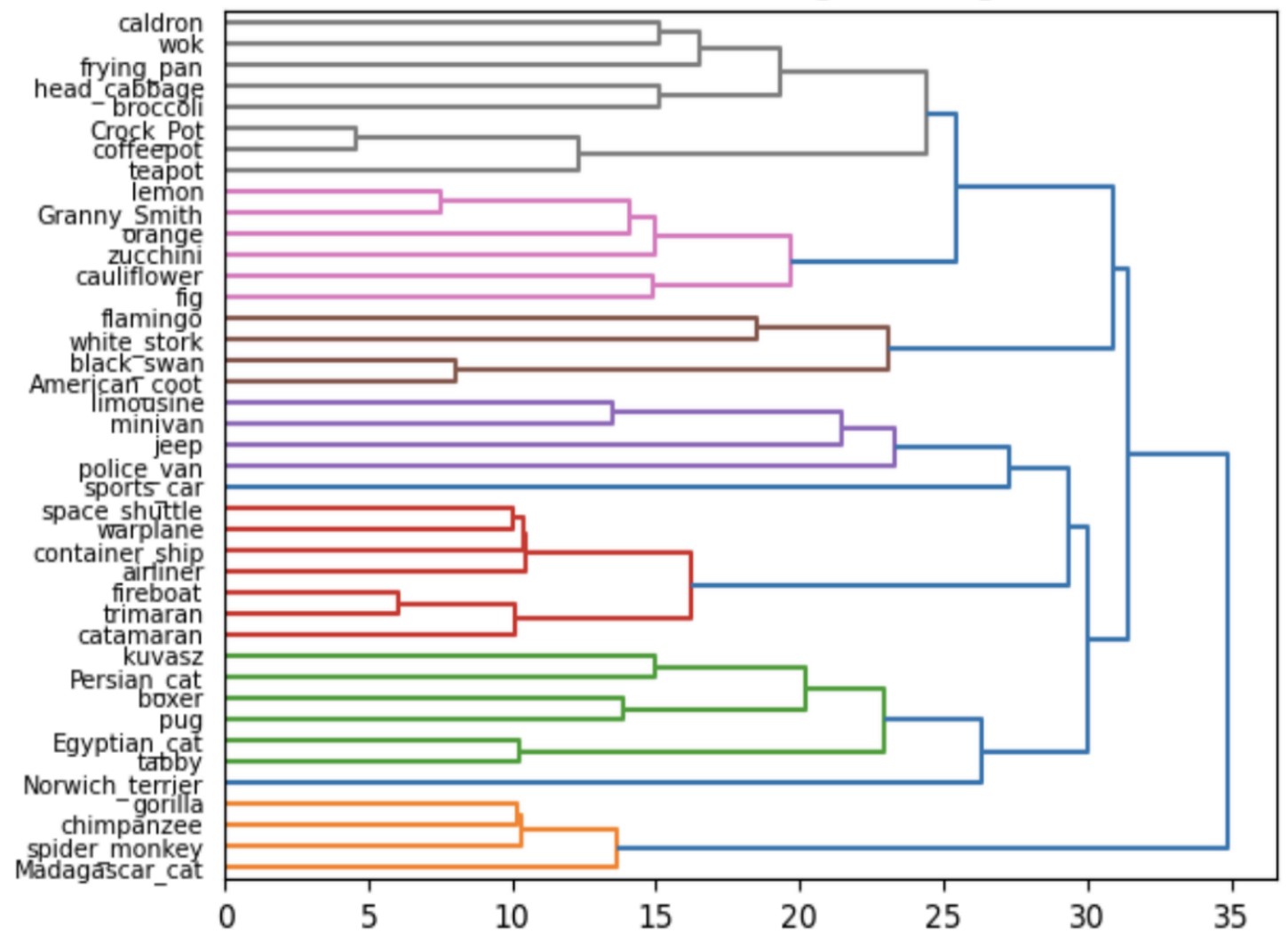}
    \caption{Visual explanation of EfficientNet on ImageNet.}
    \label{fig:EfficientNetImgNet}
\end{figure}

\begin{figure}[ht]
    \centering
    \includegraphics[width=\linewidth]{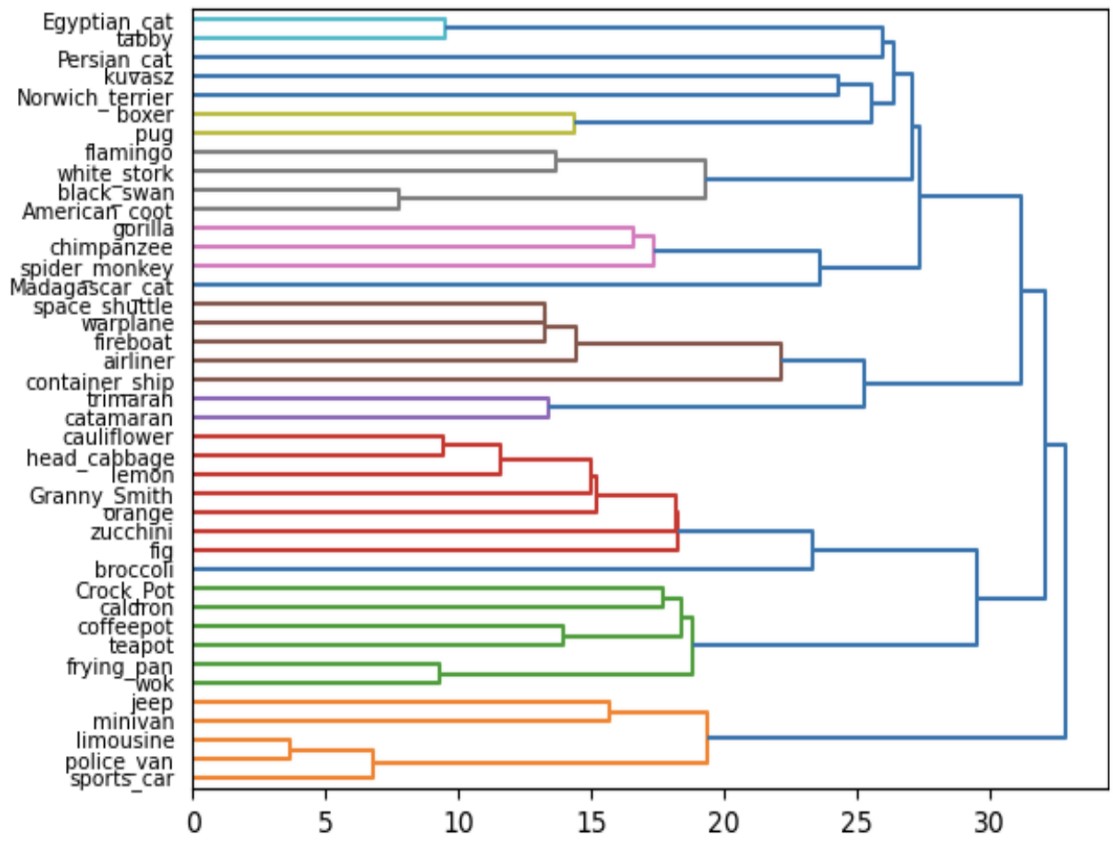}
    \caption{Visual explanation of VGG on ImageNet.}
    \label{fig:VGGImgNet}
\end{figure}

\begin{figure}[ht]
    \centering
    \includegraphics[width=\linewidth]{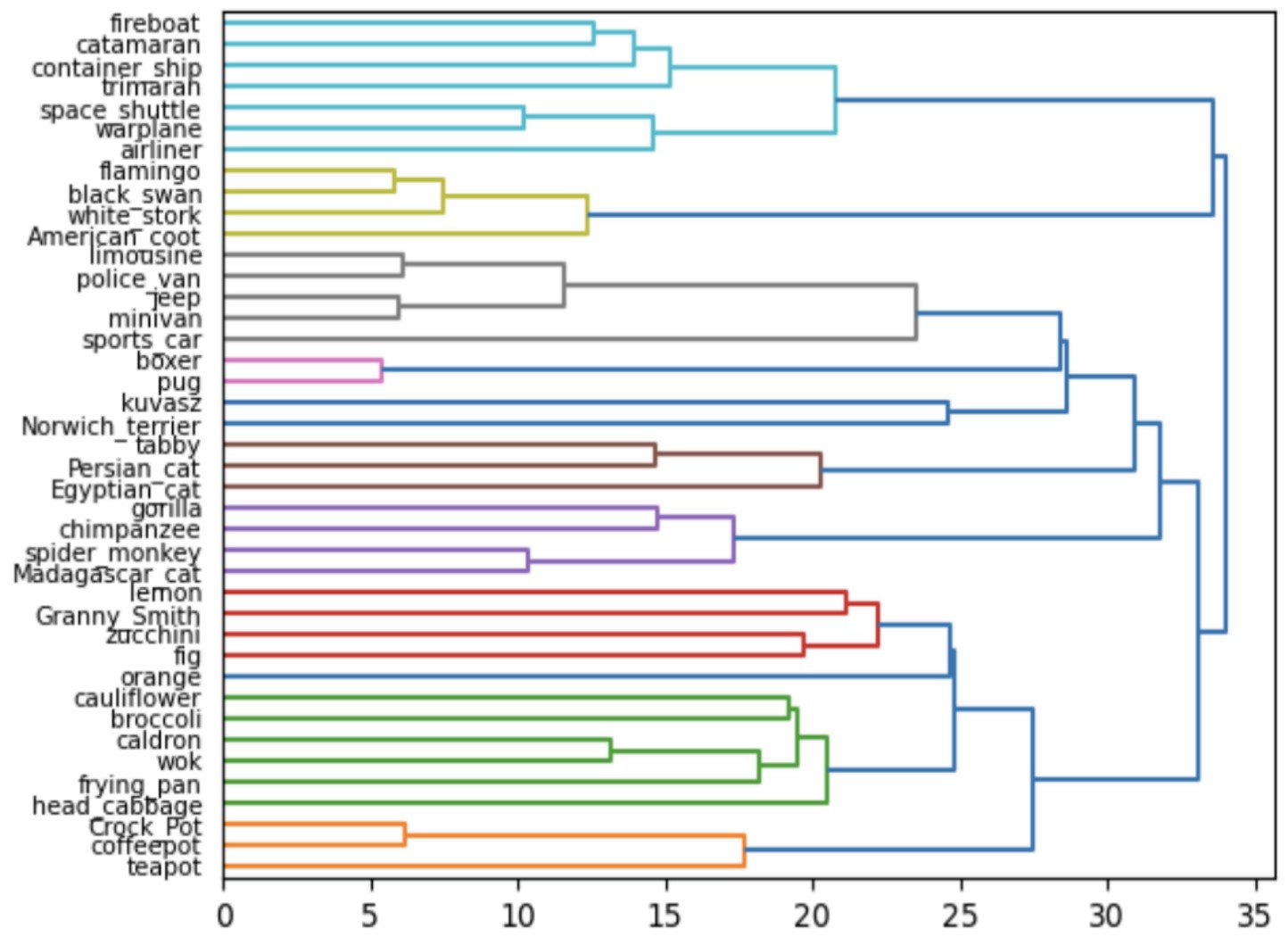}
    \caption{Visual explanation of MobileNet on ImageNet.}
    \label{fig:MobileNetV2}
\end{figure}

The resulting visual explanation for each experimented NN on ImageNet test set is presented in Figures~\ref{fig:ResNetImgNet},~\ref{fig:EfficientNetImgNet},~\ref{fig:VGGImgNet} and~\ref{fig:MobileNetV2}.  

\commentout{
\begin{figure*}[ht]
    \centering
    \includegraphics[width=0.7\textwidth]{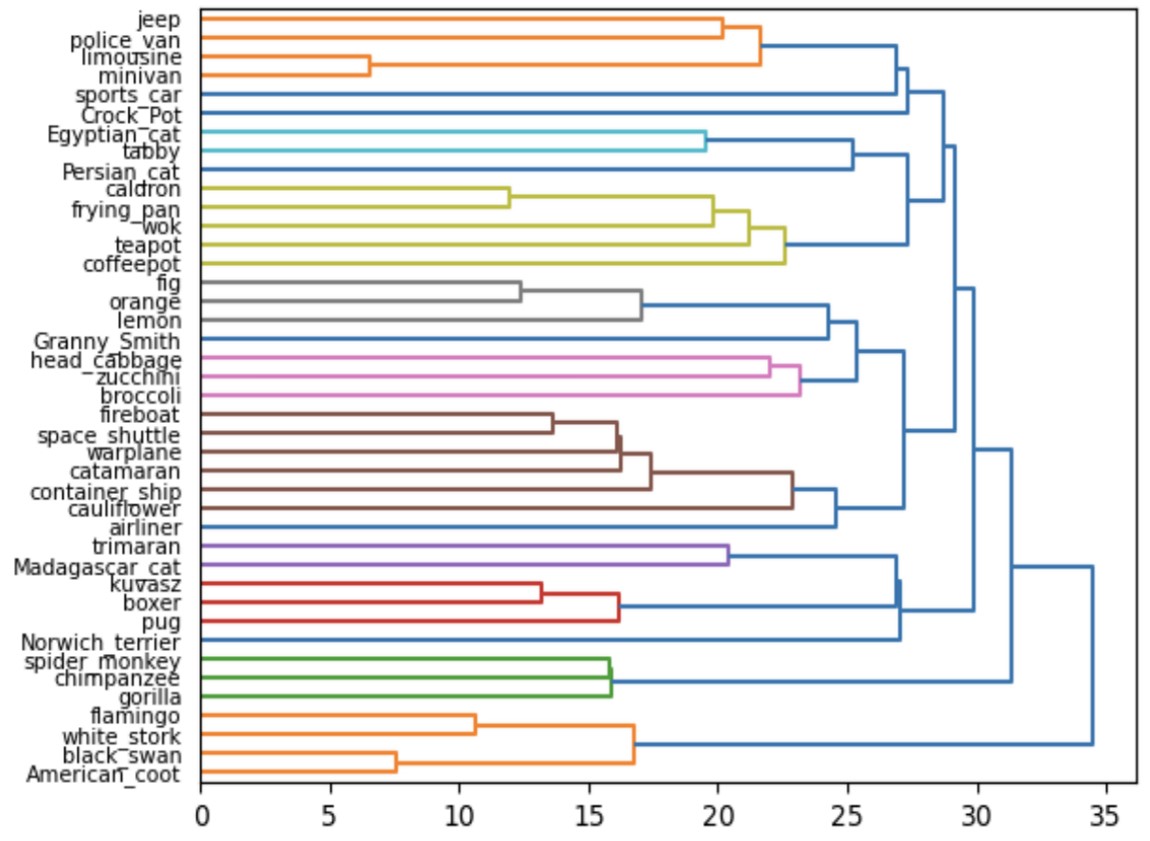}
    \caption{Visual explanation of GoogLeNet on ImageNet.}
    \label{fig:GoogLeNetImgNet}
\end{figure*}

\begin{figure*}[t]
    \centering
    \includegraphics[width=0.7\textwidth]{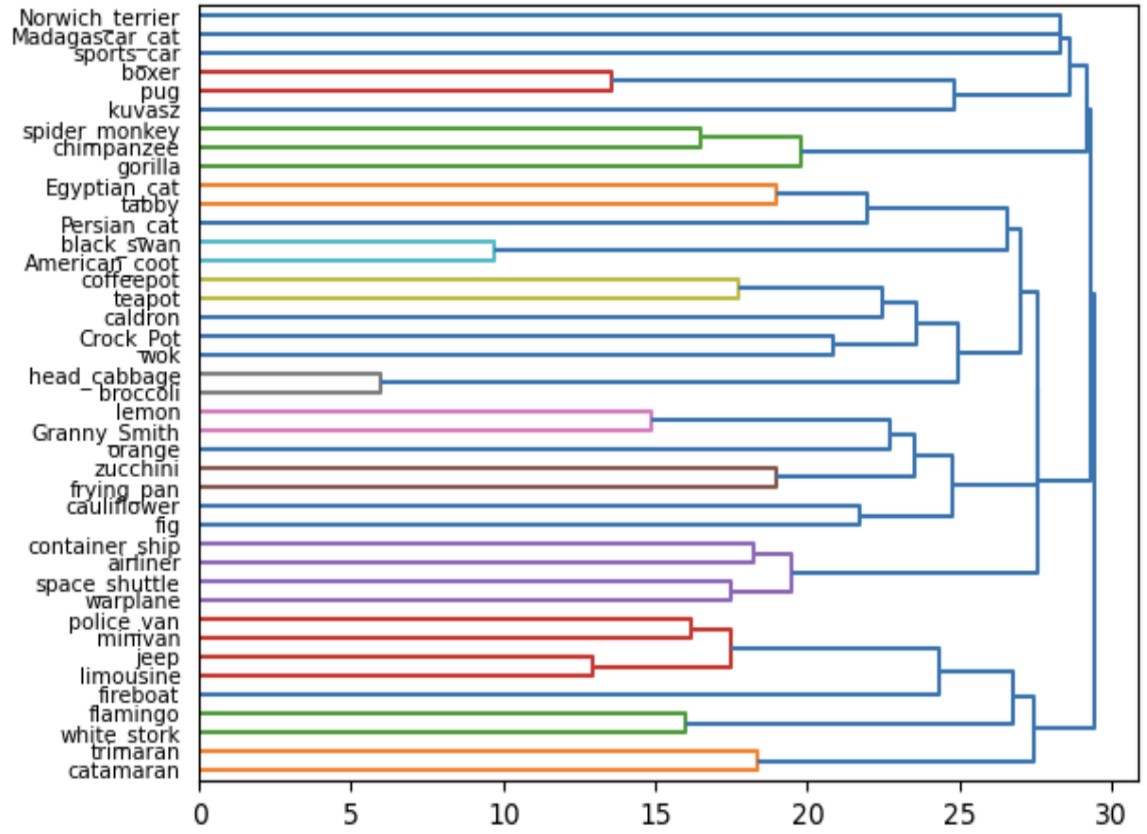}
    \caption{Visual explanation of NasNetLarge on ImageNet.}
    \label{fig:NasNetLargeImgNet}
\end{figure*}
}

\subsection{Visual Explanations Comparative Discussion}\label{ss:visual} 
Figure~\ref{fig:ResNetImgNet} shows that for ResNet at distance nearly 15 the following clusters are formed: \{warplane, airliner, space shuttle\}, \{minivan, police van\},\{container ship, fireboat\},\{trimaran, catamaran\}, \{jeep,limousine\}, which represent the different concepts of 'air transport', 'vans', 'sea transport', 'watercraft', 'land transport', respectively. These are then clustered together (with sports car) to a more general concept represented by 'vehicles' below distance 25. In addition, at distance just above 15 the following clusters are formed: \{chimpanzee, gorilla, Madagascar cat, spider monkey\}, \{Egyptian cat, Persian cat, tabby\}, \{pug, boxer, kuvasz\}, which represent the different concepts of 'primates', 'cats' and 'dogs', respectively. These are then clustered together (with Norwich terrier) to a more general concept represented by 'land animals' just above distance 25. 

Similarly, at distance nearly 15 the following clusters are formed: \{wok, frying pan\}, \{coffeepot, teapot\} (the concepts 'pans' and 'pots'), which are then clustered (with Crock pot and caldron) to a more general concept represented by 'kitchen utensils'. In addition, at distance nearly 15 the following clusters are formed: \{broccoli, head cabbage\}, \{Granny Smith, zucchini\}, \{cauliflower, lemon, fig\}, then clustered (with orange) to a more general concept represented by 'fruits and vegetables' at distance nearly 25. Finally, at distance nearly 15 the following clusters are formed: \{white stork, flamingo\}, \{black swan, American coot\}, ('long-leg water-birds' and 'pedal-leg water-birds', respectively), which are then clustered together above distance 25 to a more general concept 'water birds'. 

Interestingly, 'water birds' group is then clustered with the 'fruits and vegetables', then with the 'kitchen utensils', only then with 'land animals' and finally with the 'vehicles'. Thus, an intuitive score for explainability relative to a human explanation will give Resnet points for clustering water birds together but deduct points for missing the animal category. Apart from the 'water birds' case, which a human would normally consider as 'animals' and less close to 'kitchen utensils', it is apparent that ResNet is capable of forming abstract concepts recognized by humans. 
Note that vegetables and kitchen utensils are reasonably clustered together as vegetables are found in the kitchen and inside pots or pans.

Figure~\ref{fig:EfficientNetImgNet} shows that for EfficientNet the clusters are formed differently. At distance nearly 15 the following clusters are formed: \{Madagascar cat, spider monkey, chimpanzee, gorilla\}, \{Egyptian cat, tabby\}, \{pug, boxer\}, \{Persian cat, kuvasz\}, which represent the different concepts of 'primates', 'cats', 'dogs' and 'hairy pets', respectively. Note that, Persian cat and kuvasz are clustered together though the first is a cat and the second is a dog. It is apparent that EfficientNet is incapable of well recognizing the different concepts 'cats' and 'dogs', as done by ResNet. 

Moreover, at distance nearly 15 the last three clusters of 'cats', 'dogs' and 'hairy pets' are grouped together (with Norwich terrier) to form the more general concept represented by 'cats and dogs'. Interestingly, this cluster is \emph{not} clustered with the 'primates' until the final cluster of all objects is formed. It is apparent that EfficientNet is incapable of well recognizing the abstract concept 'land animals', as done by ResNet. In addition, the clusters: \{catamaran, trimaran, fireboat\}, \{airliner, container ship, warplane, space shuttle\}, \{minivan, limousine\} are formed at distance nearly 15. Note that the watercraft container ship is clustered at this distance with air transport vehicles probably due to its large size. Only at distance almost 30 these clusters are grouped together (with jeep, police van and then with sports car) to the more general concept of 'vehicles'.  

Similarly, at distance nearly 15 the following clusters are formed:  \{American coot, black swan\}, \{white stork, flamingo\}, which are then clustered together to a more general concept represented by 'water birds' just below distance 25. Additionally, at distance 15 the clusters:  \{fig, cauliflower\}, \{lemon, Granny Smith, orange, zucchini\} are formed. These are then clustered together to the more general concept  represented by 'fruits and vegetables' at distance 20. Note that the cluster \{broccoli, head cabbage\} is also formed at distance 15, however, it is then clustered with \{caldron, wok\} and frying pan at distance nearly 20 and then with \{Crock pot, coffeepot\} clustered with teapot, at distance just below 25 forming a cluster that may be represented as 'round kitchen items'. This cluster is formed probably due to a image of a broccoli inside a wok. Nevertheless, Resnet did not link these labels while seeing the same image. 

It is apparent that EfficientNet is incapable of well recognizing the abstract concept plants ('fruits and vegetables'), as done by ResNet, and rather relies on visual resemblance (possibly based on characteristics like roundness or hairiness). Thus, although EfficientNet claims to be more efficient by reaching similar accuracy with less parameters, its "efficiency" does not come without a cost since it is unable to 'perceive' more complex concepts. Therefore, even though it may correctly classify a test set, an adversarial image might easily cause it to confuse one object with another.

Figure~\ref{fig:VGGImgNet} shows that for VGG the clusters are formed such that 'land vehicles' as well as 'kitchen utensils' are grouped together. In addition, VGG is capable to create the general concept of 'fruits and vegetables', as done by ResNet, as well as the concept of 'air and sea transport'. Similarly, it is cable to create the concepts 'water birds', 'dogs' and 'cats'. It is apparent VGG is capable to form the abstract \emph{separate} concepts of 'cats' and 'dogs' even better than ResNet that could not add the Norwich terrier to the 'dogs'. 

Moreover, VGG is capable to create the concept represented by 'primates'. This cluster is then grouped together with the 'cats and dogs' and the 'water birds' clusters to form the more general abstract concept of 'animals'. Thus, it is apparent that VGG is capable to recognize the abstract concept of 'animals', which ResNet could not recognize (as ResNet separated 'water birds' from the other 'land animals' and grouped them first with 'kitchen utensils'). Note, however, that VGG failed to recognize the abstract concept of 'vehicles', as ResNet is capable, since VGG separated the 'air and sea transport' cluster from the 'land vehicles' and grouped them first with 'animals'.

Figure~\ref{fig:MobileNetV2} shows that for MobileNet the clusters are differently formed. Here, a single cluster representing the general concept 'kitchen utensils' is created. Yet, it is apparent that MobileNet failed to create different concepts for 'kitchen utensils' and 'fruits and vegetables', as done in ResNet or VGG. Similarly, MobileNet creates clusters, which can be represented by the concepts 'primates', 'cats', 'dogs', 'land vehicles', 'water birds', 'air transport' and 'sea transport'. In addition, 'air transport' and 'sea transport' are grouped to a single 'air and sea transport' cluster. 

Nevertheless, it is apparent that MobileNet is incapable of well recognizing the abstract concepts of neither 'vehicles' (which is recognized by by ResNet) nor 'animals' (which is recognized by VGG). Note that all 'dogs' are first grouped with the 'land vehicles', then with 'cats' and then with 'primates' at distance just above 30 forming a concept represented by 'land objects'. Similarly, between distance 30 and 35 the clusters of 'water birds' and 'air and sea transport' are grouped together to a single cluster that can be represented by the abstract concept 'air and sea objects'.

\begin{figure}[ht]
    \centering
    \includegraphics[width=0.8\linewidth]{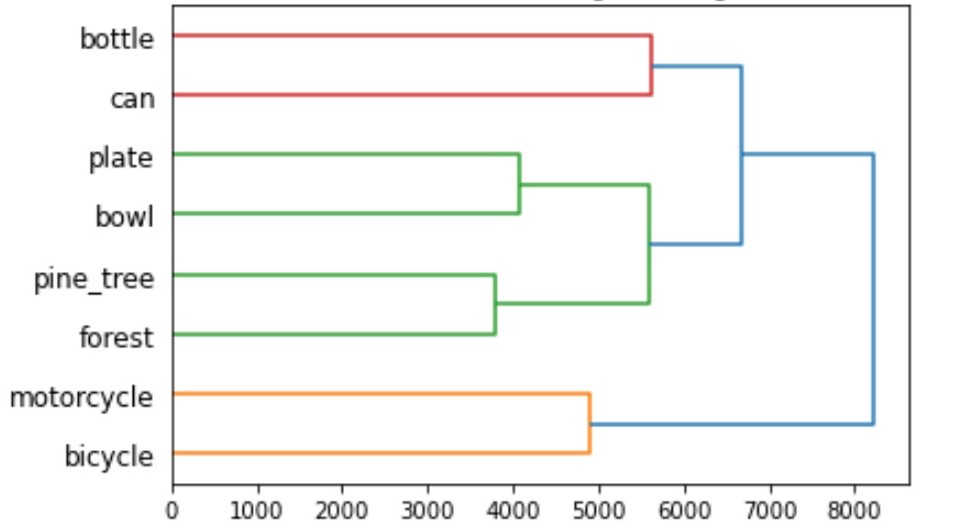}
    \caption{Visual explanation of ResNet on CIFAR100.}
    \label{fig:Cifar}
\end{figure}

\paragraph{Results on the CIFAR100 Dataset.} We retrained the chosen NNs on CIFAR100 dataset using transfer learning from ImageNet.\footnote{Due to pages limit we only give here the ResNet results.}
Figure~\ref{fig:Cifar} presents the resulting visual explanation generated for ResNet on a demonstrative sample from CIFAR100. It is apparent that ResNet is capable of creating some abstract concepts, however, seems  less successful than on ImageNet. Note that CIFAR100 has a very low resolution images of $32\times32$ pixels, thus, NNs trained on it can more easily mistake between labels. Nevertheless, Figure~\ref{fig:Cifar} demonstrates that our approach is robust enough to give an expressive visual explanation under these conditions as well.

\subsection{Query Verbal Explanations Results}\label{ss:verbal}
Table~\ref{tab:Exp} gives the results of the query verbal explanation scores presented in Equations~\ref{eq:score1} and~\ref{eq:score2} (see Section~\ref{sec:method}). The results numerically support the qualitative analysis of the visual explanations presented in Subsection~\ref{ss:visual}, demonstrating a similar trend. In particular, ResNet explanatory measurement outperforms the other less complex models in the total score, and achieves relatively high score in sub-groups queries. Conversely, MobileNet which is simpler in every aspect than both ResNet and EfficientNet, achieves inferior explanatory measurement in the total score as well as in sub-groups queries scores. This result is in-line with our a-priori assumption that more complex models will present better concept generation ability. Surprisingly, the relatively shallow VGG, which has a huge learnt parameter set, presents scores competitive to ResNet (except for the vehicles sub-group queries score) and outperforms the deeper MobileNet. 
\begin{table}[t]
\centering
\resizebox{.95\columnwidth}{!}{
\begin{tabular}{|c|c|c|c|c|c|}
\hline
    - & Total & Animals & Vehicles & Plants & Utensils\\
\hline
    ResNet50 & \textbf{0.89} & 0.9 & 0.98 & \textbf{0.75} & \textbf{1}\\
    EfficientNetB0 & 0.61 & 0.27 & \textbf{1} & 0.7 & 0.5\\
    VGG16 & 0.82 & \textbf{1} & 0.6 & \textbf{0.75} & \textbf{1}\\
    MobileNetV2 & 0.52 & 0.5 & 0.51 & 0.62 & 0.48\\
\hline
\end{tabular}}
\caption{Explanatory measurement scores on ImageNet}
\label{tab:Exp}
\end{table}
\section{Conclusions and Future Work}
In this paper we propose a novel XAI approach based on NMA and hierarchical clustering. We experimented our approach on several known NNs that vary in size and shape. The experiments results lead to the following conclusions:
\begin{enumerate}
    \item Though all NNs were queried on the same set of images, their behaviour visual explanations varied and reflected a different hierarchy for the different experimented NNs. This fact demonstrates the usability of our approach to reflect NNs latent process of concepts generation. 
    \item All experimented NNs demonstrated capability to form abstract concepts to some extent. However, there is an apparent difference between the quality of such abstract concept generation among the different NNs. As expected, the more complex ResNet was capable of forming high-quality abstract hierarchical concepts superior to the less complex EfficientNet and MobileNet. 
    \item Interestingly, the relatively shallow VGG, that has a huge learnt parameters-set size, was capable of forming high-quality abstract hierarchical concepts, competing with the deep ResNet and outperforming the deeper EfficientNet and MobileNet.
    It is commonly assumed that lower-layers of NNs seek visual features such as edge or triangular detection and upper fully-connected layers organize all features into a concept \cite{molnar2022,DBLP:conf/cvpr/BauZKO017}.
    Furthermore, it is commonly assumed that fully-connected layers require many parameters with no additional gain and that deeper CNN architectures are preferred. Hence, modern CNN architectures have very thin to almost none fully-connected layers \cite{DBLP:conf/icml/TanL19}. Our results seem to challenge this assumption. 
\end{enumerate}

\commentout{In future work, we intend to more thoroughly examine the effect of parameters-set size vs.~the network depth on the ability of NNs to create abstract concepts. We also want to test the stability of the proposed approach results under various size and content test-sets.} 
In future work, we intend to perform white-box testing as well as examine the suggested approach effect on adversarial attacks construction and prevention. We believe that NMA analysis with hierarchical clustering can support not only 'black box' explanations, as demonstrated in this paper, but rather be used within 'white box' dissection methods by examining a given NN layer by layer in order to identify where new concepts are formed and in which neurons. In addition, NMA may also enhance adversarial attack strategies by pinpointing which labels are closely related within the same concepts, thus, aid developing countermeasures against them. We, therefore, believe that the suggested approach is a promising new path for XAI research to follow.

\newpage
\bibliography{AAAI25}
\end{document}